\documentclass{article} 
\usepackage{sp_conf}
\usepackage{graphicx}
\usepackage{times}
\ninept

\newcommand{\appears}[4]{%
  \begin{picture}(0,0)
  \put(0,0){\raisebox{#1}{%
    \hspace*{#2}%
    \parbox[t]{#3}{\normalsize\tt #4}}}
  \end{picture}}

\def\for{`\textsf{\footnotesize FOR}'}
\def\you{`\textsf{\footnotesize YOU}'}
\def\church{`\textsf{\footnotesize CHURCH}'}
\def\of{`\textsf{\footnotesize OF}'}
\def\foryou{`\textsf{\footnotesize FOR YOU}'}
\def\ofchurch{`\textsf{\footnotesize OF CHURCH}'}

\def\hub4e{\textsl{Hub--4E}}
\def\drf{\hat{f}}
\def\event{{\cal E}}
\def\vocab{{\cal V}}
\def\eg{{\em e.g.\/}}

\def\ie{{\em i.e.\/}}

\title{Variable Word Rate N-Grams}
\name{Yoshihiko Gotoh \hspace{0.5in} Steve Renals
	\thanks{This work was funded by UK EPSRC grant GR/M36717.}}
\address{University of Sheffield, Department of Computer Science \\
	Regent Court, 211 Portobello St., Sheffield S1 4DP, UK \\
	e-mail: \{y.gotoh, s.renals\}@dcs.shef.ac.uk}

\begin{document}
\maketitle
\appears{2.4in}{-0.2in}{6.9in}{
	Proceedings of the 25th International Conference on Acoustics, Speech,
	and Signal Processing (ICASSP-2000), Istanbul, June 2000}
\vspace{-0.3in}

\begin{abstract}
  The rate of occurrence of words is not uniform but varies from document to
  document.
  Despite this observation, parameters for conventional $n$-gram language
  models are usually derived using the assumption of a constant word rate.
  In this paper we investigate the use of variable word rate assumption,
  modelled by a Poisson distribution or a continuous mixture of Poissons.
  We present an approach to estimating the relative frequencies of words or
  $n$-grams taking prior information of their occurrences into account.
  Discounting and smoothing schemes are also considered.
  Using the Broadcast News task, the approach demonstrates a reduction of
  perplexity up to 10\%.
\end{abstract}

\section{Introduction} 
\label{sc: introduction}

In both spoken and written language, word occurrences are not random but vary
greatly from document to document.
Indeed, the field of information retrieval (IR) relies on the degree of
departure from randomness as a discriminative indicator.
IR systems are typically based on unigram statistics (often referred to as
a ``bag-of-words'' model), coupled with sophisticated term weighting schemes
and similarity measures~\cite{sparck:rep98}.
In an attempt to mathematically realise the intuition that an occurrence of
a certain word may increase the chance that the same word is observed later,
several probabilistic models of word occurrence have been proposed.
Much of this work has evolved around the use of (a mixture of) the Poisson
distribution~\cite{bookstein:art74,harter:art75a,robertson:sigir94}.
Recently, Church and Gale have demonstrated that a continuous mixture of
Poisson distributions can produce accurate estimates of variable word
rate~\cite{church:art95}.
Lowe has introduced a beta-binomial mixture model which was applied to topic
tracking and detection~\cite{lowe:euro_sp99}.

Although a constant word rate is an unlikely premise, it is nevertheless
adopted in many areas including $n$-gram language modelling.
In order to address the problem of variable word rate, several adaptive
language modelling approaches have been proposed with a moderate degree of
success.
Typically, some notion of ``topic'' is inferred from the text according to the
``bag-of-words'' model.
Information from different language model statistics (\eg, a general model
and/or models specific to each topic) are then combined using methods such as
mixture modelling~\cite{kneser:icassp93} or maximum
entropy~\cite{rosenfeld:csl96}.
The \emph{dynamic cache model\/}~\cite{kuhn:pami90} is a related approach,
based on an observation that recently appearing words are more likely to
re-appear than those predicted by a static $n$-gram model.
It blends cached unigram statistics for recent words with the baseline
$n$-grams using an interpolation scheme.

Theoretically, it should not be necessary to rely on an \emph{ad hoc\/} device
such as a cache in order to model variable word occurrences.
All the parameters of a language model may be completely determined according
to probabilistic model of word rate, such as a Poisson mixture.

In this paper, we outline the theoretical background for modelling the
variable word rate, and illustrate a key observation that word rates are not
static using spoken data transcripts.
The constant word rate assumption is then eliminated, and we introduce
a variable word rate $n$-gram language model.
An approach to estimating relative frequencies using prior information of word
occurrences is presented.
It is integrated with standard $n$-gram modelling that naturally involves
discounting and smoothing schemes for practical use.
Using the DARPA/NIST \hub4e North American Broadcast News task, the approach
demonstrates the reduction of perplexity up to 10\%.

\section{Modelling Variable Word Rates} 
\label{sc: word}

In this section, we illustrate how the assumption of a constant word rate
fails to capture the statistics of word occurrence in spoken (or written)
documents.
We show that the word rate is variable and may be modelled using a Poisson
distribution or a continuous mixture of Poissons.

\subsection{Poisson Model} 

The Poisson distribution is one of the most commonly observed distributions in
both natural and social environments.
It is fundamental to the queueing theory: under certain conditions, the number
of occurrences of a certain event during a given period, or in a specified
region of space, follows a Poisson distribution
(a \emph{Poisson process\/}~\cite{karlin:book75}).

By assuming randomness in a Poisson process, word rate is no longer uniform.
Firstly, we provide a loose definition of a document as a unit of spoken (or
written) data of a certain length that contains some topic(s), or content(s).
We consider a model in which a word occurs at random in a fixed length
document.
For a set of documents we assume that each document produces this word
independently and that the underlying process is the Poisson with a single
parameter $\lambda > 0$.

Formally, a Poisson distribution is a discrete distribution (of a random
variable $X$) which is defined for $x = 0,1,\cdots$ such that
\begin{eqnarray}
  \theta^{[p]}(x)
	= {\cal P}(X = x;\ \lambda)
	= \frac{e^{- \lambda} \lambda^x}{x!}
\label{eq: poisson}
\end{eqnarray}
whose expectation and variance are given by $E[X] = \lambda$ and
$V[X] = \lambda$, respectively~\cite{degroot:book70}.

\subsection{Poisson Mixture --- Negative Binomial Model} 

A less constrained model of variable word rate is offered by a multiple of
Poissons, rather than a single Poisson.

\begin{figure}
\centerline{\includegraphics[width=3.2in]{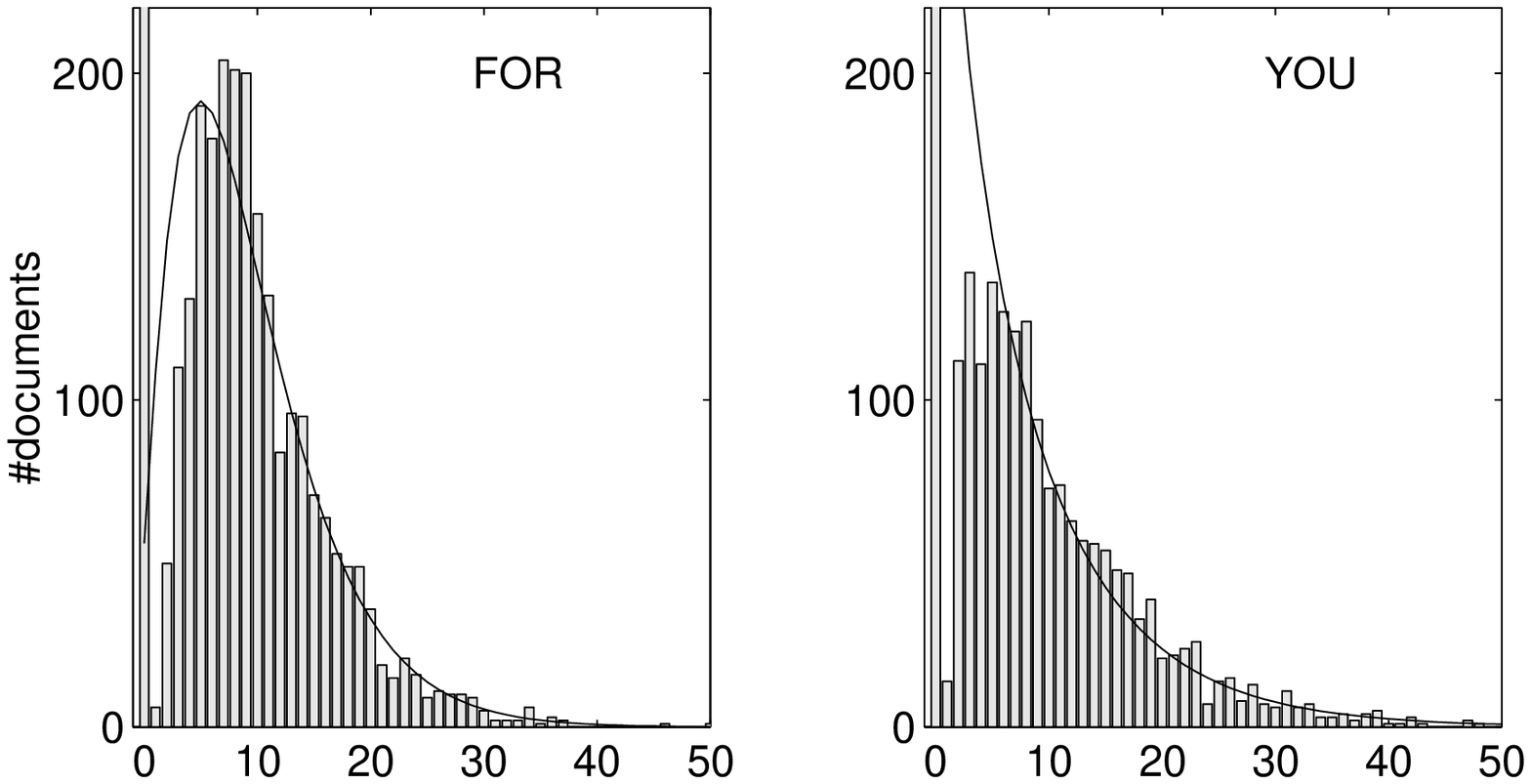}}
\smallskip
\centerline{\includegraphics[width=3.2in]{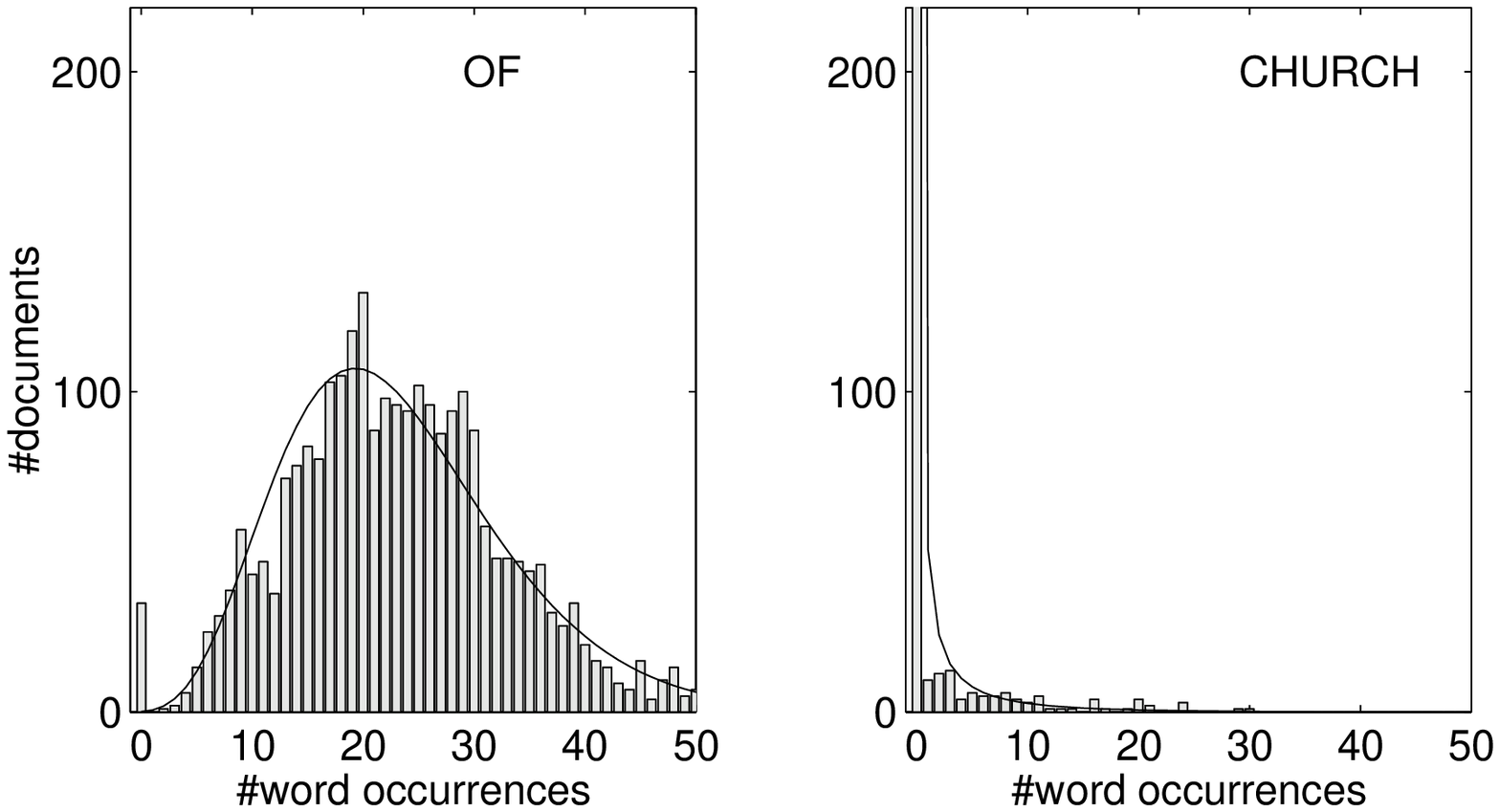}}
\caption{The occurrence of words (unigrams) varies between documents.
	Histograms show the number of word occurrences for \for, \you, \of,
	and \church\ from a set of 2583 documents, each containing at least
	100 words (average: 497 words).
	A negative binomial distribution (solid line) was used to approximate
	each histogram.
	The number of word occurrences were normalised to 1000-word length
	documents.}
\label{fig: uhist}
\end{figure}

Suppose the parameter $\lambda$ of the pdf~(\ref{eq: poisson}) is distributed
according to some function $\phi(\lambda)$, then we define a continuous
mixture of Poisson distributions by
\begin{eqnarray}
  \theta(x) = \int_0^\infty \theta^{[p]}(x) \phi(\lambda) d\lambda \ .
\label{eq: mixture}
\end{eqnarray}
In particular, if $\phi(\lambda)$ is a gamma distribution, \ie,
\begin{eqnarray}
  \phi(\lambda) = {\cal G}(\lambda;\ \alpha, \beta)
	= \frac{\lambda^{\alpha - 1} e^{- \frac{\lambda}{\beta}}}
	{\beta^\alpha \Gamma(\alpha)}
\label{eq: gamma}
\end{eqnarray}
for $\alpha > 0$ and $\beta > 0$, then the integral~(\ref{eq: mixture}) is
reduced to a discrete distribution for $x = 0,1,\cdots$ such that
\begin{eqnarray}
  \theta^{[nb]}(x) \hspace{-0.1in} &=& \hspace{-0.1in}
	{\cal NB}(X = x;\ \alpha, \beta)
\nonumber \\
  &=& \hspace{-0.1in}
	\left( \begin{array}{c} \alpha + x - 1 \\ x \end{array} \right)
	\frac{\beta^x}{(1 + \beta)^{\alpha + x}} \ .
\label{eq: nb}
\end{eqnarray}
This $\theta^{[nb]}(x)$ is a negative binomial distribution\footnote{
	Let $\phi(\lambda)$ be ${\cal G}(\lambda;\ \alpha, \beta)$
	in~(\ref{eq: mixture}).
	This integration is straightforward using the definition of the gamma
	function, $\displaystyle
	\Gamma(\alpha) = \int_0^\infty t^{\alpha - 1} e^{-t} dt$, and the
	recursion, $\Gamma(\alpha + 1) = \alpha \Gamma(\alpha) $.
	The resultant pdf~(\ref{eq: nb}) has a slightly unconventional form in
	comparison to that in most of standard textbooks
	(\eg,~\cite{degroot:book70}), but is identical by setting a new
	parameter $\displaystyle \gamma = \frac{1}{1 + \beta}$ with
	$0 < \gamma < 1$.}
and its expectation and variance are respectively given by $E[X] = \alpha\beta$
and $V[X] = \alpha\beta (\beta + 1)$.

\subsection{Word Occurrences in Documents} 

The histograms in figure~\ref{fig: uhist} show the number of word (unigram)
occurrences in spoken news broadcast, taken from transcripts of the \hub4e
Broadcast News acoustic training data (1996--97).
These transcripts were separated into documents according to section markers
and those with less than 100 words were removed, resulting 2583 documents
containing slightly less than 1.3 million words in total.
In the following, the number of word occurrences were normalised to 1000-word
length documents.

\for\ and \you\ appeared approximately the same number of times across all the
transcripts.
Using a constant word rate assumption, they would have been assigned
a probability of around 0.0086.
However their occurrence rates varied from document to document; about 11\%
and 33\% of all documents did not contain \for\ and \you\ (respectively),
while 1\% and 3\% contained these words more than 30 times.
This seems to indicate that occurrences of \for\ is less dependent on the
content of the document.
A negative binomial distribution was used to model the variable word rate in
each case (the solid line in figure~\ref{fig: uhist}).

The negative binomial seems to model word occurrence rate relatively well for
most vocabulary items, regardless of frequency.
Figure~\ref{fig: uhist} illustrates this for one of the most frequent words
\of\ (probability of 0.023 according to the constant word rate assumption) and
the less frequently occurring \church\ (less than 0.00029).
In particular, \church\ appeared only in 93 out of 2583 documents, but 28 of
them contained more than 10 instances, suggesting strong correlation with
document content.

\begin{figure}
\centerline{\includegraphics[width=3.2in]{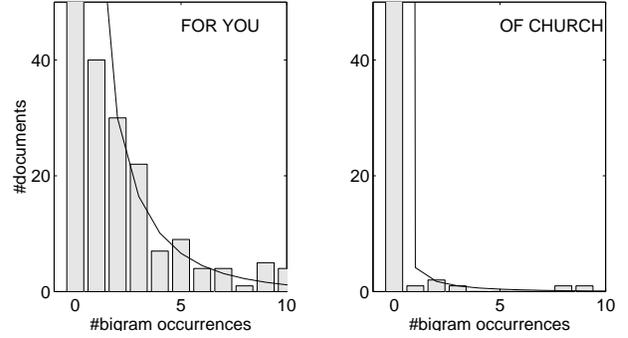}}
\caption{Variable bigram occurrence rates.
	Histograms show the number of bigram occurrences (in normalised
	1000-word documents) for \foryou\ and \ofchurch, combinations of
	unigrams used in figure~\ref{fig: uhist}.
	They are fitted by negative binomial distributions (solid lines).}
\label{fig: bhist}
\end{figure}

We also collected statistics of bigrams appearing in the Broadcast News
transcripts.
Figure~\ref{fig: bhist} show histograms and their negative binomial fits for
bigrams \foryou\ and \ofchurch.
Although very sparse (\eg, they appeared in 127 and 6 documents, respectively),
this suggests that variable bigram rate can also be modelled using
a continuous mixture of Poissons.

\section{Variable Word Rate Language Models} 
\label{sc: lm}

Taking word occurrence rate into account changes a probabilistic language
model from a situation akin to playing a lottery, to something closer to
betting on a horse race:
the odds for a certain word improve if it has come up in the past.
In this section, we eliminate the constant word rate assumption and present
a variable word rate $n$-gram language model.

\subsection{Relative Frequencies with Prior Word Occurrences} 

Let $f(w\geq n_w)$ denote a relative frequency after we observe $n_w$
occurrences of word $w$.
It is calculated by
\begin{eqnarray}
  f(w\geq n_w) = \frac{1}{N}
	\frac{\displaystyle m_w - \sum_{j=0}^{n_w - 1} j\cdot \theta_w(j)}
	{\displaystyle 1 - \sum_{j=0}^{n_w - 1} \theta_w(j)} \ .
\label{eq: rf_var}
\end{eqnarray}
The function is defined for $n_w = 0,1,\cdots,N$, where $N$ is a fixed
document length (\eg, $N$ is normalised to 1000 in figures~\ref{fig: uhist}
and~\ref{fig: bhist}).
$\theta_w(j)$ is the occurrence rate for word $w$ in an $N$-length document
(\eg, Poisson, negative binomial), satisfying
\begin{displaymath}
  \sum_{j=0}^N j\cdot \theta_w(j) = m_w \ , \hspace{0.2in}
	\displaystyle \sum_{j=0}^N \theta_w(j) = 1 \ .
\end{displaymath}
In particular,
\begin{eqnarray}
  \displaystyle f(w\geq 0) = \frac{m_w}{N} \ ,
\label{eq: rf_const}
\end{eqnarray}
which corresponds to the case with no prior information of word occurrence.
For the conventional approach with the constant word rate assumption,
this $f(w\geq 0)$ is not modified regardless of any word occurrences.
Further, function~(\ref{eq: rf_var}) satisfies our intuition; the value of
$f(w\geq n_w)$ increases monotonically as the number of observation $n_w$
accumulates (easy to verify), and it reaches a unity (`1') when $n_w = N$.

\begin{figure}
\centerline{\includegraphics[width=3.2in]{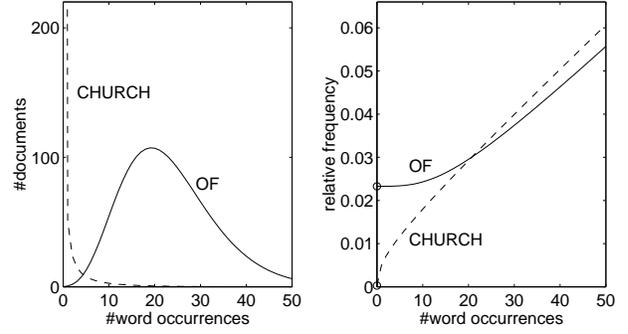}}
\caption{The left figure shows word occurrence rates for \of\ and \church\ in
	documents of (normalised to) 1000-word length, modelled by negative
	binomial distributions (identical to those in figure~\ref{fig: uhist}).
	The right figure demonstrates relative frequencies after a certain
	number of word occurrences.
	Circles (`\textsf{o}') correspond to relative frequencies under the
	constant word rate assumption
	(0.023 for \of\ and 0.00029 for \church).}
\label{fig: rfrequency}
\end{figure}

The characteristics of function~(\ref{eq: rf_var}) are illustrated in
figure~\ref{fig: rfrequency}.
The right hand figure shows relative frequencies for \of\ and \church\ after
a certain number of previous observations of the word.
It indicates that the first few instances of the frequent word (\of) do not
modify its relative frequency very much, but have a substantial effect on the
relative frequency of the less common word (\church).
As the number of observations increases, the former is caught up by the latter.

Finally, in order to convert this relative frequency model to any type of
probabilistic model for language, normalisation is required.
This is achieved by dividing $f(w\geq n_w)$ by
$\displaystyle \sum_{w\in\vocab} f(w\geq n_w)$, where $\vocab$ implies a set
of vocabulary.
Variable relative frequencies for bigrams can also be calculated in a similar
fashion.

\subsection{Discounting and Smoothing Techniques} 

For any practical application, smoothing of the probability estimates is
essential to avoid zero probabilities for events that were not observed in
the training data.
Let $\event(w | v)$ denote a bigram entry (a word $v$ followed by $w$) in the
model.
Further, $f(w | v\geq n_{w | v})$ implies a relative frequency after we
observe $n_{w | v}$ occurrences of the bigram.
A bigram probability $p(w | v\geq n_{w | v})$ may be smoothed with a unigram
probability $p(w\geq n_w)$.
Using the interpolation method~\cite{jelinek:proc80}:
\begin{eqnarray}
  p(w | v\geq n_{w | v}) \hspace{-0.1in}
	&=& \hspace{-0.1in} \drf(w | v\geq n_{w | v})
\nonumber \\
	&& \hspace{-0.1in} +\ \{1 - \alpha(v)\} \cdot p(w\geq n_w)
\label{eq: interpolation}
\end{eqnarray}
where $\drf(w | v\geq n_{w | v})$ implies a ``discounted'' relative frequency
(described later) and
\begin{eqnarray}
  \alpha(v) = \sum_{w\in\event(w | v)} \drf(w | v\geq n_{w | v})
\label{eq: nonzero}
\end{eqnarray}
is a non-zero probability estimate (\ie, the probability that a bigram entry
$\event(w | v)$ exists in the model).
Alternatively, the back-off smoothing~\cite{katz:assp87} may be applied:
\begin{eqnarray}
  p(w | v\geq n_{w | v}) = \left\{\begin{array}{l}
		\drf(w | v\geq n_{w | v}) \hspace{4.0mm}
		\mbox{if } \event(w | v) \mbox{ exists}, \\
		\beta(v)\cdot p(w\geq n_w) \hspace{4.0mm} \mbox{otherwise}.
		\end{array}\right.
\label{eq: backoff}
\end{eqnarray}
In~(\ref{eq: backoff}), $\beta(v)$ is a back-off factor and is calculated by
\begin{eqnarray}
  \beta(v) = \displaystyle \frac{1 - \alpha(v)}
	{\displaystyle 1 - \sum_{w\in\event(w | v)} \drf(w\geq n_w)} \ .
\label{eq: factor}
\end{eqnarray}
A unigram probability $p(w\geq n_w)$ can be obtained similarly by smoothing
with some constant value.

Finally, a number of standard discounting methods exist for constant word rate
models (see, \eg,~\cite{katz:assp87,ney:pami95}).
Analogous discounting functions for variable word rate models may be
\begin{eqnarray}
  \drf_{abs}(w | v\geq n_{w | v}) = f(w | v\geq n_{w | v}) - \frac{c}{N}
\label{eq: absolute}
\end{eqnarray}
for the absolute discounting, and
\begin{eqnarray}
  \drf_{gt}(w | v\geq n_{w | v}) = d\cdot f(w | v\geq n_{w | v})
\label{eq: gt}
\end{eqnarray}
for the Good-Turing discounting.
Discounting factors ($c$ and $d$) may be obtained using zero prior information
case --- \ie, $f(w | v\geq 0)$'s of all bigrams in the model --- and the rest
should be referred to, \eg,~\cite{katz:assp87} or~\cite{ney:pami95}.

\subsection{Language Model Perplexities} 

\begin{figure}
\centerline{\includegraphics[width=3.2in]{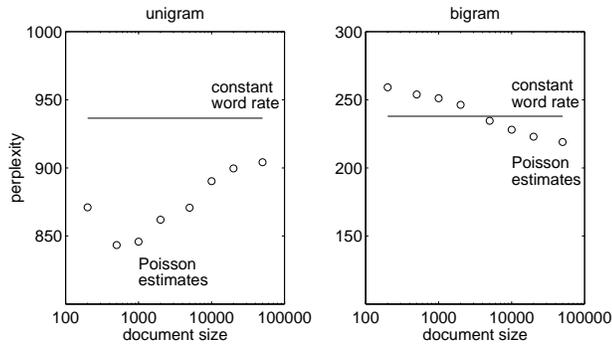}}
\caption{Unigram and bigram perplexities for the reference (key) transcription
	of 1997 \hub4e evaluation data.
	Conventional models (constant word rate) are compared with models
	using Poisson estimates of variable word rate.
	Document length for the latter was normalised to between 200 and
	50\,000.}
\label{fig: perplex}
\end{figure}

As noted in section~\ref{sc: word}, we extracted 2583 documents from the
transcripts of the Broadcast News acoustic training data, each with a minimum
of 100 words.
A vocabulary of 19\,885 words was selected and 390\,000 bigrams were counted.
In these experiments, the absolute discounting scheme~(\ref{eq: absolute}) was
applied, followed by interpolation smoothing~(\ref{eq: interpolation}).
Figure~\ref{fig: perplex} shows perplexities for the reference (key)
transcription of the 1997 \hub4e evaluation data, containing three hours of
speech and approximately 32\,000 words.
Using conventional modelling with a constant word rate assumption, unigram
and bigram perplexities were 936.5 and 237.9, respectively.

For the variable word rate models, the Poisson distribution was adopted
because of simplicity in calculation.
The number of word occurrences were normalised to $N$-word length document
with $N$ being between 200 and 50\,000, and the model parameters were modified
`on-line' during the perplexity calculation.
For each occurrence of a word (bigram) in the evaluation data, a histogram of
the past $N$ words (bigrams) was collected and their relative frequencies were
modified according to the Poisson estimates (appropriate normalisation
applied), then discounted and smoothed.

As figure~\ref{fig: perplex} indicates, the variable word rate models were
able to reduced perplexities from the constant word rate models.
A unigram perplexity of 843.4 (10\% reduction) was achieved when
$N = 500$, and a bigram perplexity of 219.0 (8\% reduction) when $N = 50\,000$.
The difference was predictable because bigrams were orders of magnitude more
sparse than unigrams.

\section{Conclusion} 
\label{sc: conclusion}

In this paper, we have presented a variable word/$n$-gram rate language model,
based upon an approach to estimating relative frequencies using prior
information of word occurrences.
Poisson and negative binomial models were used to approximate word occurrences
in documents of fixed length.
Using the Broadcast News task, the approach demonstrated a reduction of
perplexity up to 10\%, indicating potential although the technique is still
premature.
Because of the data sparsity problem, it is not clear if the approach can be
applied to language model components of current state-of-the-art speech
recognition systems that typically use 3/4-grams.
However, we believe this technique does have application to problems in the
area of information extraction.
In particular, we are planning to apply these methods to the named entity
annotation task, along with further theoretical development.


\end{document}